\documentclass{article}
\usepackage{graphicx}
\usepackage{xcolor}
\usepackage{mathtools}
\usepackage{geometry}
\usepackage{booktabs}
\usepackage{multirow}
\usepackage{siunitx}
\usepackage{caption}
\usepackage{subcaption}
\usepackage{soul}
\usepackage{algorithmic}
\usepackage{algorithm}
\usepackage{url}
\usepackage{pdfpages}
\usepackage{amsfonts}
\usepackage{array}
\usepackage{amsmath}

\DeclareMathOperator*{\argmin}{arg\,min}
\usepackage[normalem]{ulem} 
\usepackage[hidelinks]{hyperref}
\usepackage[flushleft]{threeparttable}
\date{}
\title{A Virtual Reality-based Training and Assessment System for Bridge Inspectors with an Assitant Drone}

\author{%
Yu Li, Muhammad Monjurul Karim, Ruwen Qin
\thanks{Y. Li, M.M. Karim, R. Qin are with the Department of Civil Engineering, Stony Brook University, Stony Brook, NY, 11794, USA.}\thanks{Corresponding author, ruwen.qin@stonybrook.edu, 2424 Computer Science Building, Stony Brook University, Stony Brook, NY 11794-4424}}

\begin{document}

\title{A Virtual Reality-based Training and Assessment System for Bridge Inspectors with an Assistant Drone}
\author{%
Yu Li, Muhammad Monjurul Karim, Ruwen Qin
\thanks{Y. Li, M.M. Karim, R. Qin are with the Department of Civil Engineering, Stony Brook University, Stony Brook, NY, 11794, USA.}\thanks{Corresponding author, ruwen.qin@stonybrook.edu, 2424 Computer Science Building, Stony Brook University, Stony Brook, NY 11794-4424}}

\maketitle
\section*{Abstract}
Over 600,000 bridges in the U.S. must be inspected every two years to identify flaws, defects, or potential problems that may need follow-up maintenance. Bridge inspection has adopted unmanned aerial vehicles (or drones) for improving safety, efficiency, and cost-effectiveness. Although drones can operate in an autonomous mode, keeping inspectors in the loop is critical for complex tasks in bridge inspection. Therefore, inspectors need to develop the skill and confidence to operate drones in their jobs. This paper presents the design and development of a virtual reality-based training and assessment system for inspectors assisted by a drone in bridge inspection. The system is composed of four integrated modules: a simulated bridge inspection developed in Unity, an interface that allows a trainee to operate the drone in simulation using a remote controller, data monitoring and analysis to provide real-time, in-task feedback to trainees to assist their learning, and a post-study assessment supporting personalized training. The paper also conducts a proof-of-concept pilot study to illustrate the functionality of this system. The study demonstrated that TASBID, as a tool for the early-stage training, can objectively identify the training needs of individuals in detail and, further, help them develop the skill and confidence in collaborating with a drone in bridge inspection. The system has built a modeling and analysis platform for exploring advanced solutions to the human-drone cooperative inspection of civil infrastructure.

\hfill\break%
\noindent\textit{Keywords}: Virtual Reality, Infrastructure Inspection, Unmanned Aerial Vehicle, Training, Performance Assessment, Sensing, Human-in-the-loop

\section{Introduction}

The U.S. Highway Bridge Inventory has approximately 617,000 bridges. 42\% of them are over 50-years old, and 7.5\% are structurally deficient \cite{ASCE2021}. To avoid catastrophic incidents, all bridges are required to be inspected every two years for identifying flaws, defects, or potential problems that may need follow-up maintenance. Traditional bridge inspection may require closing the traffic and the use of heavy equipment such as a snooper truck. Inspecting a bridge needs a crew of inspectors working at the site for many hours. Some field operations are dangerous, such as climbing up to high bridge columns. 

To make bridge inspection safer, faster,  cheaper, and less interruptive to the traffic, unmanned aerial vehicles or drones have been adopted for use. A drone can conveniently access various locations of a bridge to capture a large amount of inspection data efficiently using sensors that it carries, such as RGB cameras and infrared cameras. Bridge inspectors can collect bridge inspection data using a drone if they have a license issued by the Federal Aviation Administration (FAA) and a waiver of the regulation ``keeping the drone within visual line-of-sight" in FAA’s Small UAS (Part 107) Regulations. Then, inspectors will bring the data back to their offices and analyze the data with the assistance of machine learning algorithms \cite{karim2021semi}. The use of drones for data collection also minimizes the traffic closure and the use of heavy, expensive equipment. A survey conducted by the American Association of State Highway and Transportation Officials (AASHTO) shows that using drones for bridge inspection can reduce the cost by 74\% \cite{AASHTOdrone2019}. Besides bridges, other low-accessible infrastructures have also been adopting drones for inspection, such as dams and penstocks \cite{ozaslan2015inspection}, transmission lines \cite{li2020two}, and railways \cite{banic2019intelligent}.

Current studies on the bridge inspection with an assistant drone mainly focus on the drone technology  (e.g., \cite{metni2007uav}) and data analysis using image processing and computer vision (e.g., \cite{karim2021semi}). The human factors aspect is largely ignored. The use of drones for bridge inspection is not to eliminate bridge inspectors but to augment their ability \cite{hopcroft2006unmanned}. Although a drone can fly in the autonomous mode by following a pre-planned path, keeping human-in-the-loop will enhance the safety, efficiency, and effectiveness of bridge inspection. There are various situations that the inspector needs to disengage the autonomous mode and take control of the drone. For example, if the inspector identifies a severe concern with a certain spot of the bridge during the inspection, the inspector can temporarily pause the autonomous mode to collect desired data around that spot. After that, the autonomous mode can be resumed to continue the planned inspection. In response to an alarming situation or a suddenly emerging need anticipated by the inspector or an Artificial Intelligence (AI) model, the inspector may also have to take control of the drone \cite{kontitsis2003uav}.

Training is essential to help inspectors gain and retain the skill and confidence in inspecting bridges with an assistant drone. Training inspectors to collect data using an assistant drone should take place progressively in multiple stages. Like aviation training systems or driving simulators, a Virtual Reality (VR)-based training and assessment system is a cost-effective tool for the early-stage training. After that, inspectors can move to the Augmented Reality (AR)-based training that either uses a virtual drone at a real inspection site or a real drone in a virtual inspection scene. Ultimately, the training will be at a real bridge site with a real drone. VR-based training systems have been developed for civil engineers in pipe maintenance \cite{shi2020impact}, bridge construction process \cite{sampaio2013virtual}, and bridge crane operation \cite{dong2009research}. Some commercial drone flight simulators have been developed as well. For example, AeroSim Drone Simulator \cite{AeroSim} offers training scenarios of inspecting wind turbines, power lines, towers, and solar panels. Moud et al. \cite{moud2019flight} also developed a first-ever drone flight simulator for construction sites. To our best knowledge, no simulator has been developed for inspector-drone cooperative bridge inspection, nor a data-driven framework for assessing the training performance of bridge inspectors.

A dedicated simulator is required for training bridge inspectors to operate a drone in their jobs. Commercial drone simulators are not tailored toward the need for training bridge inspectors, according to our discussions with researchers in human factors and a center for training inspectors. Some unique features of bridge inspection differentiate it from other types of inspection, such as the traffic passing the bridge, complex and diverse structures, narrow irregular spaces between structural elements such as diaphragms and interlayers. Moreover, factors that may impact the drone-assisted bridge inspection are broad, including job site-related, drone-related, task-related, and human-related factors. But commercial drone simulators only considered some of those, such as the wind speed and direction, battery level, and task difficulty level (e.g., \cite{AeroSim, moud2019flight, gabriel2016workload, rauffet2016eye}). The skill and confidence developed using simulators for inspecting other types of infrastructure are not transferred to the bridge inspection effectively.

A dedicated assessment method for training bridge inspectors is desired too. The assessment utilizes the data captured from the simulation training to measure inspectors' performance using specially-designed metrics. The stakeholder of the training system usually would like to set a baseline for rating inspectors' overall performance in utilizing an assistant drone as excellent, good, acceptable, and others, for example. Without well-defined performance metrics, the baseline is difficult to determine objectively. Feedback to inspectors, both in-training and post-training, helps accelerate their learning processes. Effective feedback to an inspector should be built on measurements of the inspector's tasks performance and human states (e.g., cognitive load, physical load, emotion, and other psychological states). The measurements are from multiple dimensions, including time, quality, productivity, safety, cost, and others. Commercial drone simulators do not include a module that provides desired measures and metrics for monitoring and assessing bridge inspectors in training. While task performance measurements have been studied widely in operations management \cite{neely1995performance}, many metrics are output-based, such as the completion time and productivity, not applicable to providing in-task feedback. Evaluation of task performance and human states can be performed by subjects themselves, peers, or evaluators. But subjective judgment lacks reliability, timeliness, and accuracy. Despite these limitations, subjective evaluation is still often used due to the low cost and the ease of implementation. The NASA Task Load Index (NASA-TLX) \cite{hart1988development} is a questionnaire commonly used by pilots for reporting their physical demand, time pressure, effort, performance, mental demand, and frustration level  \cite{fidopiastis2009impact, gabriel2016workload}.

Filling gaps in the literature, contributions of this paper are twofold:
\begin{itemize}
    \item the design and development of a VR-based system devoted to the training and assessment of bridge inspectors assisted by a drone (TASBID), and
    \item a data-driven method with unique measures and metrics for analyzing and understanding inspectors' needs for training and assistance.
\end{itemize}
The source code of TASBID is publicly available for download at Github \cite{TASBID_Github}. The remainder of this paper is organized as the following. The next section presents the proposed system. Section \ref{sec:Illustration} exhibits the functionality of the system using a small-scale pilot study. In the end, Section \ref{sec:Conclusion} concludes the study and summarizes directions of future work. 

\section{The Training and Assessment System}
\label{sec:System}
\begin{figure*}[htbp]
    \centering
    \includegraphics[scale=0.5]{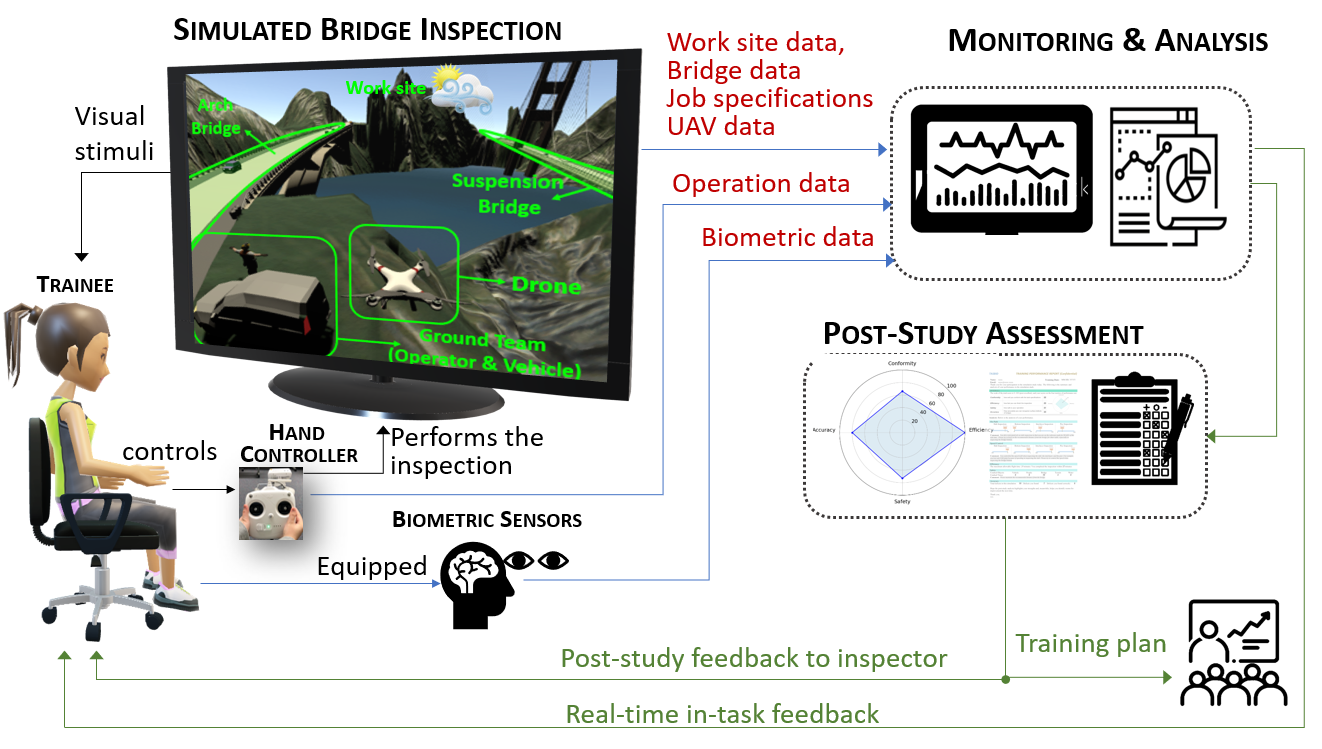}
    \caption{Architecture of the training and assessment system for bridge inspection with an assistant drone (TASBID)}
    \label{fig:OV-1}
\end{figure*}

The architecture of the training and assessment system for bridge inspectors with an assistant drone (TASBID) is illustrated in Fig.\ref{fig:OV-1}. The system is designed to consist of four modules: the bridge inspection simulation, an interface that allows the bridge inspector (the trainee) to operate the drone in the simulated inspection, monitoring \& data analysis, and the post-study assessment. The trainee, who could be equipped with biometric sensors, operates the drone in the simulated inspection using a remote controller. Streaming data of the inspector and the drone are monitored to provide real-time in-task feedback to the inspector. The data on job specifications, the bridge, and the site are references for monitoring and analysis. Upon completing a training, a comprehensive assessment based on the collected data is performed to provide both the required information for designing an individualized training plan and the post-study feedback to the trainee. Below is a discussion of the four modules in detail.

\subsection{Simulated Bridge Inspection}
The inspection simulation created in Unity is illustrated in Fig.\ref{fig:OV-1}. The simulation provides the trainee with the visual stimulus of drone-assisted bridge inspection. To assure it is close to the real-world work context, the simulation is designed to include five major elements of bridge inspection: the ground team, the drone, bridges, the job site, and example tasks.

\subsubsection{The Ground Team}
The ground team in the simulation comprises an inspector (the virtual counterpart of the trainee) and a truck. The simulation defines the location of the ground team where the drone takes off and returns to. The simulation provides two views side by side on the screen to the trainee. The left is the inspector's view at the site, and the right is the camera's view from the drone. During the simulation training, the trainee can switch her/his gazes between the two views to focus on the most useful one. For example, the inspector's view can be useful during taking-off and landing. The camera's view is what the trainee would concentrate on during the inspection and for the visual navigation.  

\subsubsection{The Drone}
The simulation adopts a drone simulation model from a Unity drone controller asset \cite{DroneController} and revises it for the bridge inspection. Parameters for modeling the drone include the drone model, mass and load, movement force, maximum forward speed, maximum sideward speed, rotation speed, slow-down time, movement sound, propellers' rotation, battery capacity, and movement types. In this simulation, eight types of movement are sufficient for the bridge inspection. They are forward \& backward, right-sideward \& left-sideward, up \& down, and right-rotation \& left-rotation. The trainee controls the movement, rotation, and speed of the drone using the remote controller. The battery level is a dynamic constraint for the drone operation, which drops gradually during the inspection. This simulation does not include the return-to-home function that can bring the drone to the home point when the battery level drops to a pre-specified level. Instead, the battery level is displayed for examining the trainee's time stress. The drone has a snapshot function that the inspector can straightforwardly use to label an event on the timeline of the inspection video. Later, the inspector can retrieve and review the labeled frames. The snapshot function is also a simple way of confirming the inspector's visual attention to an area of concern.

\subsubsection{Bridges}
The simulation uses a Unity asset named Road Architect \cite{RoadArchitect} to create the bridge models, wherein multiple types of bridges are available for choice and redesign. The simulation includes an arch bridge and a suspension bridge to provide trainees with different experiences in training. For example, the arch bridge in the simulation has cramped spaces where controlling the drone is challenging to the trainee. Road Architect defines the structural elements of the bridges. Accordingly, the spatial-temporal relationship between the drone and specific bridge elements during the simulated inspection can be determined. Defects such as cracks are added to the surface of some bridge elements to assess the trainee's situational awareness during the inspection. 

\subsubsection{The Job Site}
Simulation of the bridge inspection site focuses on creating the geographic context, the environmental condition, and the traffic condition at the bridges. Bridges to be inspected sit on a lake in a mountain area. Bridge inspection needs to be conducted in the daytime with clear weather although a sudden change in the weather may occur in rare cases. Therefore, only wind under level five of the Beaufort Wind Scale has been considered as a possible weather impact in the system. The wind factor is simulated by adding the force value and direction in Unity. Since most commercial drones can be flown in the wind between 10 and 30 mph, TASBID considers three levels of wind: light, gentle, and medium. They correspond to the wind speed around 2mph, 11mph and 22mph, and cause the force of 0.12N, 3N and 12N, respectively. A three-dimensional vector can set up the wind direction. Lighting condition is another common factor impacting the inspection. To create a more realistic lighting condition, the simulation turns on the Global Illustration in Unity to simulate the light reflected from the water surface. Some dark areas of bridges, such as the bridge bottom, are still present although the natural lighting is good. TASBID is designed to include tasks that inspect dark areas of bridges. The drone in the simulation is equipped with a light. Trainees can turn it on or off according to their needs by pressing ``B" on the keyboard. Traffic volume is modeled as well because inspectors may feel pressure when flying the drone near the traffic. Vehicles moving on the bridges are included in TASBID using a free traffic simulation asset \cite{traffsimulation}. The total number of vehicles can be increased or decreased as desired.

\subsubsection{Inspection Tasks}
Tasks selected for the simulation training must capture representative scenarios of the real-world inspection. TASBID includes four tasks that have various shapes of inspection paths (e.g., a long straight line vs. multiple short lines, and a large curve vs. a small circle), types of accessible space (spacious and narrow), lighting conditions (bright vs. dark), and various levels of complexity in controlling the drone (gross vs. fine control, and movement vs. rotation control). 
\begin{itemize}
    \item Task 1 is to inspect the slab of the arch bridge from one side. The accessible space is spacious, and the lighting condition usually is not a concern. The recommended inspection path for this task is a straight line along one side of the bridge. The trainee controls the drone and moves it along the slab from one end to another end of the bridge. 
    \item Task 2 is to inspect the bridge bottom. The accessible space is spacious, but the lighting condition might not be ideal. The trainee needs to adjust the drone’s position frequently when moving it along the arch-shaped bridge bottom.
    \item Task 3 is to inspect the interlayer of the arch bridge in a narrow space, and the lighting condition might not be ideal. The trainee needs to delicately control the movement and rotation of the drone to capture both the upper side and the down side of the interlayer area safely.
    \item Task 4 is to inspect the corrosion situation of the suspension bridge at a pier. The path for the drone is a circle with a small radius near the water surface. The trainee needs to rotate the drone when moving around the pier frequently. 
\end{itemize}

The task sequence presented above is just a recommendation. The trainee can plan and decide the sequence of tasks. 

\subsection{Interface between the Trainee and the Drone}
The trainee operates the drone in the simulated inspection using a remote controller. Currently, TASBID uses a Phantom 2 DJI controller for this purpose. The controller is connected to Unity using the vJoy device driver \cite{vjoystick} and the method in mDjiController \cite{mDjiController}. The trainee adjusts the joysticks of the controller to control the movement, rotation, and speed of the drone. The trainee's operations of the drone are recorded as time series data.

\subsection{Monitoring \& Data Analysis}
TASBID can collect six types of data from the study, as Fig.\ref{fig:OV-1} illustrates. The work site characteristics, the bridge models, the drone model, and job specifications are pre-specified data that do not change during a study. The flight data of the drone and the trainee's operation data are the frame-level streaming data that vary in each time of the study.

\subsubsection{Streaming Data}
A simulated inspection is captured by a sequence of $N$ frames, indexed by $i$. Given the fixed frame rate, $f$, the total duration of an inspection is $N/f$. The starting frame is defined as the time when the drone is taking off. The ending frame corresponds to the time when the drone lands near the ground team, or it cannot continue to finish the inspection (e.g., the battery drains or the drone crashes into the traffic), whichever occurs the first. Let $O_i$ and $D_i$ denote the trainee's operation data and the drone flight data, respectively, collected at any frame $i$. 

The trainee operates the remote controller that has four-axis inputs for controlling the movement, rotation, and speed of the drone. Besides, the trainee can press ``B" on the keyboard to turn on/off the light and ``P" to take ``snapshots" during the inspection. Therefore, the trainee's operation data are time series data in six dimensions:
\begin{equation}
O_{i}=[o_{fb, i}, o_{rl,i}, o_{ud,i}, o_{rt,i}, o_{b,i}, o_{p,i}]
\end{equation}
where
\begin{itemize}
    \item[] $o_{fb, i}$: Forward (+) $\&$ Backward (-),
    \item[] $o_{rl, i}$: Right (+) $\&$ Left (-) Sideward,
    \item[] $o_{ud, i}$: Up (+) $\&$ Down (-),
    \item[] $o_{rt, i}$: Right (+) $\&$ Left (-) Rotation,
    \item[] $o_{b, i}$: Turning on (1) $\&$ off (0) the light,
    \item[] $o_{p, i}$: Taking a snapshot (1) $\&$ not (0).
\end{itemize}
    
The drone flight data include the position, velocity, and the remaining batter level of the drone:
\begin{equation}
D_{i}=[\vec{L}_i, v_i, b_i]=[l_{x,i}, l_{y,i}, l_{z,i}, v_i, b_i],
\end{equation}
where $\vec{L}_i=(l_{x,i}, l_{y,i}, l_{z,i})$ are the 3D coordinates of the drone's location in the earth reference system, $v_i$ is the linear speed of drone, and $b_i$ is the remaining battery level in percentage.

\subsubsection{On-path Analysis}
Although the drone has the gimbal and zoom functions to make the data collection more flexible, the flexibility is bounded. Therefore, the inspector’s ability to send the assistant drone to suitable locations is still critical to obtaining desired inspection data in desired quality.  For the training purpose, TASBID recommends reference paths appropriate for performing individual tasks, but not for the entire job, to the trainee. Let $t$ be the index of tasks and $n$ be the index of reference points. $\{\vec{p}_{t,n}|n=1, \dots, N_t\}$ defines the reference flying path for the drone in task $t$. 

Denote $X_{t,i}$ as the binary variable indicating if the drone in frame $i$ is on the reference path of task $t$, for any $t$ and $i$. $\sum_{t=1}^TX_{t,i}\leq 1$ for any $i$, indicating the drone cannot be on more than one task simultaneously. Using Algorithm \ref{algorithm_inpath} below, the analysis module evaluates if the drone is on the reference path for task $t$. Specifically, the algorithm uses the reference path of task $t$ and the location of the drone as inputs to determine the value of the binary variable $X_{t,i}$. 

Using the outputs of Algorithm \ref{algorithm_inpath}, the starting frame of task $t$, $I_{t,s}$, and the ending frame, $I_{t,e}$, are determined accordingly. The analysis module treats the first frame when $X_{t,i}$ is one as the starting frame for task $t$ and the last frame when $X_{t,i}$ is one as the ending frame:
\begin{equation}
\begin{aligned}
    &I_{t,s}=\min_{i}\{i|X_{t,i}=1\}\\
    &I_{t,e}=\max_{i}\{i|X_{t,i}=1\}
\end{aligned}
\end{equation}

\begin{algorithm}[t!]
\caption{On-path analysis for task $t$ in any frame $i$}
\begin{algorithmic}
\STATE // $\{\vec{p}_{t,n}|n=1,\dots,N_t\}$: reference points that define the reference path for the drone in task $t$,
\STATE // $\vec{L}_i$: position of the drone in frame $i$,
\STATE // $l(\vec{L}_i,\vec{p}_{t,n}$): the distance between the drone and the reference point $p_{t,n}$,
\STATE // $n^\ast$: the index of the reference point with the shortest distance to the drone,
\STATE // $v_{t,n}$: the segment of the reference path, defined by $\vec{p}_{t,n}$ and its adjacent point(s),
\STATE // $l(\vec{L}_i,v_{t,n})$: the distance from the drone to any location on the segment $v_{t,n}$,
\STATE // $l^\ast_{t,i}$: the minimum distance from the drone to the reference path,
\STATE // $\overline{l}_t$: the threshold distance for identifying if the drone is on the reference path for task $t$,
\STATE // $X_{t,i}$: binary variable indicating whether the drone in frame $i$ is on the reference path of task $t$.
\vspace{0.1in}
\STATE Step 1: find the reference point with the shortest distance to the drone, $\vec{p}_{t, n^\ast}$, where
\STATE $n^\ast:=\argmin_n \{l(\vec{L}_i,\vec{p}_{t,n})|n=1,\dots,N_t\}$.
\vspace{0.1in}
\STATE Step 2: The shortest distance from the drone to the reference path is computed as:
\STATE $l^\ast_{t,i}=\min l(L_i, v_{t,n^\ast})$
\vspace{0.1in}
\STATE Step 3: Determine if the drone in frame $i$ is on path: 
\STATE $X_{t,i}=\textbf{1}\{l^\ast\leq\overline{l}_t\}$
\vspace{0.1in}
\STATE \textbf{Return} ($X_{t,i}$, $l^\ast_{t,i}$)
\end{algorithmic}
\label{algorithm_inpath}
\end{algorithm}

\subsubsection{On-speed Analysis}
%\noindent{\it Analysis of Speed Control.}

A speed limit, $\overline{v}$, is also specified for the inspection tasks. Setting a speed limit for the drone would help lower the chance of motion blur in the inspection video data. The value of $\overline{v}$ is 10 mph in TASBID. Similarly, a binary variable, $X_{s,i}$, is defined to indicate if the drone in frame $i$ is speeding when performing inspection tasks.
\begin{equation}
    X_{s,i}=\textbf{1}\{v_i> \bar{v},i\in\cup_{t=1}^T[I_{t,s},I_{t,e}]\}.
    \label{eq:speeding}
\end{equation}

\subsubsection{Crash Analysis}
A crash in the simulation is defined as an event that the drone touches traffic agents, the bridges, the terrain, or the waterbody. The simulation can sense the type of an object the drone crashes into and track timings of crash events. $X_{h,i}$ is a binary variable indicating if the drone in frame $i$ touches a human in the traffic. $X_{v,i}$ is another binary variable indicating if the drone crashes into a vehicle in the traffic. $X_{o,i}$ is a categorical variable indicating if the drone touches any other objects. A crash event may last for multiple frames. Therefore, whenever $X_{h,i}$ turns from zero to one, the simulation identifies the occurrence of a crash into a human, indicated by a binary variable $X_h$:
\begin{equation}
X_h=\textbf{1}\{\exists X_{h,i}=1\},
\end{equation}
and another binary variable $X_v$ indicates if a crash into a vehicle happened:
\begin{equation}
X_v=\textbf{1}\{\exists X_{v,i}=1\}.
\end{equation}
Crashing into other objects will not terminate the study. At the end, the total number of crashes into other objects will be
\begin{equation}
X_o=\sum_{i=1}^N\textbf{1}\{X_{o,i}\neq 0\,\&\,X_{o,i-1}=0\}.
\end{equation}

\subsubsection{Visual Attention Analysis}
The study randomly places $X_d$ surface defects on the bridges. The trainee can take a snapshot if she/he believes an area of concern is found. The total number of snapshot events is:
\begin{equation}
    X_{pd}=\sum_{i=1}^N o_{p,i}.
\end{equation}
The snapshots may include false detection. The number of true detection is $X_{td}$.

\subsubsection{Real-time, In-task Feedback}
\begin{figure}[htbp]
    \centering
    \includegraphics[width=\columnwidth]{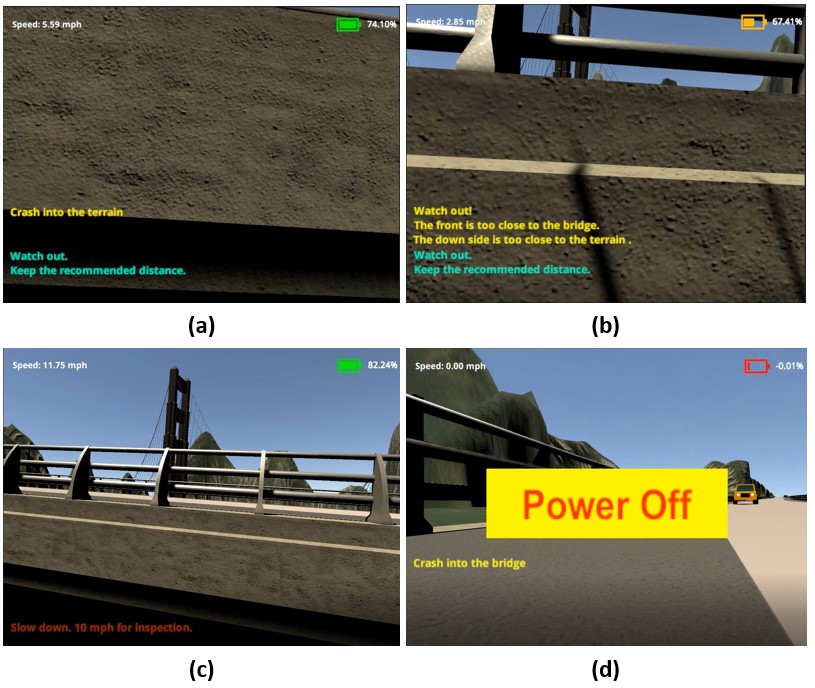}
    \caption{Illustration of real-time, in-task feedback}
    \label{fig:warning}
\end{figure}

Using the monitoring data and measurements calculated from the data, real-time, in-task feedback is provided to the trainee. To raise the trainee's attention to job safety and task specifications, TASBID provides five types of information, illustrated in Fig. \ref{fig:warning}.

The remaining battery level is updated in real-time and displayed at the upper right corner of the camera view, as Fig. \ref{fig:warning} illustrates. The battery icon is in green color when the remaining power is 70\% or higher, yellow if between 30\% and 70\%, and otherwise in red. The battery icon starts to flush once the remaining power drops below 30\%. The displayed battery level set a time constraint to encourage the trainee to finish the inspection before the drone runs out of power. The drone's speed is always displayed at the upper left corner of the camera view. Three types of messages may appear at the bottom left when certain conditions occur: 
\begin{itemize}
    \item A message about speeding will show up at the bottom left corner if $X_{s,i}$ in Equation (\ref{eq:speeding}) is one. 
    \item A message to remind the recommended distance from the bridge elements will appear if the drone is off-path, far away from the bridge element to inspect (i.e., Algorithm \ref{algorithm_inpath} returns $X_{t,i}=0$ and $l^\ast_{t,i}\leq 8$m for the inspection task $t$).
    \item A message appears if the drone senses any object within 2.5m to the center of the drone or crashes into anything (i.e., whenever $X_{h,i}$, $X_{v,i}$, or $X_{o,i}$ turns from zero to a positive value).
\end{itemize}

\subsection{Post-Study Assessment}
After a simulation training ends, data collected from the training are further used to perform a comprehensive post-study assessment. The assessment covers the trainee's task performance and self-assessment using a questionnaire. 

\subsubsection{Task Performance}
\label{subsubsec:Task Performance}
It is desired to make the bridge inspection faster, cheaper, safer, more objective, and less interruptive to the traffic. Therefore, TASBID evaluates trainees' job/tasks performance from multiple dimensions: conformity, efficiency, safety, and accuracy, which are important to the bridge inspection.

The trainee's ability to conform with task specifications is term {\it conformity}. Conformity captures inspectors’ essential ability to operate the drone along desired paths, move it to desired locations, and maintain the recommended speed, during the inspection. Conformity positively contributes to the quality of data collection. The ability to be on-path in performing task $t$ is measured by the percentage of task time when the drone is on the reference paths of the tasks:
\begin{equation}
P_{p,t} = \frac{\sum_{i=I_{t,s}}^{I_{t,e}}X_{t,i}}{I_{t,e}-I_{t,s}+1}.\\
\label{eq:path}
\end{equation}
To measure the trainee's on-speed ability in performing task $t$, a weighted sum of times when the drone is speeding is calculated, and the weights are the ratios of speed to speed limit:  
\begin{equation}
P_{s,t }=\frac{\sum_{i=I_{t,s}}^{I_{t,e}} (v_i/\bar{v})X_{s,i}}{I_{t,e}-I_{t,s}+1}.
\label{eq:overspeed}
\end{equation}
Then, the conformity is an aggregation of $P_{p,t}$ and $P_{s,t}$ for all tasks:
\begin{equation}
P_C = \omega_p \sum_{t=1}^TP_{p,t} + \omega_s \sum_{t=1}^TP_{s,t},
\label{eq:conformity}
\end{equation}
where $\omega_p$ is the gain coefficient for on-path and $\omega_s$ is the loss coefficient for speeding. The range of $P_C$ in TASBID is [-100,100]. The maximum score occurs if the drone is always on-path and never speeding in all tasks. The minimum score occurs when the drone is never on-path and always flying at its maximum speed. The maximum speed of the drone in TASBID is $30$ mph and the speed limit for inspection is $10$ mph. Therefore, $\omega_p$ and $\omega_s$ are set to be 25 and -25/3, respectively.

The trainee's ability to finish the inspection with fewer resources and less waste is termed time {\it efficiency}. It is selected as a training performance metric for encouraging inspectors to keep the inspection cost-effective. Multiple critical values are defined with respect to the time efficiency of trainees. $\underline{\tau}$ defines the cut-off point of the inspection time for receiving the highest score and $\bar{\tau}$ is the maximum allowable flight time for the drone. The battery drains if the inspection would go beyond $\bar{\tau}$. Let $X_b$ be a binary variable indicating if the drone fails to return to the ground team due to running out of power. $X_b$ equals one if $N/f>\bar{\tau}$, and zero otherwise. Accordingly, The score of time efficiency, $P_E$, is calculated as:
\begin{equation}
P_E=[\omega_{e0}+\omega_{e1}(N/f-\underline{\tau})^+](1-X_b)+\omega_bX_b.
\label{eq:efficiency}
\end{equation}
The range of $P_E$ score is [-100,100]. $\omega_{e0}$ in  Equation (\ref{eq:efficiency}) is set to be 100, representing the highest efficiency score a trainee receives if the inspection is done by $\underline{\tau}$. $\omega_b$ is set to be -100, indicating the trainee fails to complete the inspection within the maximum allowable time $\overline{\tau}$ and thus loses 100 points. $P_E$ score will be 0 if the inspection is completed at the defined maximum allowable time $\overline{\tau}$. $\omega_{e1}=-\omega_{e0}/(\overline{\tau}-\underline{\tau})$, representing the score deduction for every additional unit of time exceeding $\underline{\tau}$. In TASBID, $\overline{\tau}$ is assumed to be 25 minutes, estimated based on the maximum flight time of representative commercial lithium battery-based drones \cite{li2018routing}. $\underline{\tau}$ is set to be 15 minutes.

Job {\it safety} is the trainee’s ability to keep the drone and other traffic agents safe during the inspection. The lack of ability to keep safe in inspection is measured by the total lost score due to crashes:
\begin{equation}
   P_S=\max[\omega_hX_h+\omega_vX_v+\omega_oX_o,P'_S]
   \label{eq:safety}
\end{equation}
where $\omega_h$, $\omega_v$, and $\omega_o$ are losses from each crash into a human, a vehicle, and any other object, respectively. In TASBID, $\omega_h$ and $\omega_v$ are set to be -100, indicating a crash into a traffic agent usually has severe consequences such as a fatality or a hospitalized incident. $\omega_o$ is set to be -3, indicating the consequence of crash into other objects is more related to the drone damage. $P'_S$ is set to be -100 in TASBID, meaning that no more points will be further deducted if the cumulative loss has reached $P'_S$. Therefore, the range of $P_S$ score is [-100,0]. Limiting the loss by $P'_S$ can avoid the scenario that safety dominates other performance metrics. 

{\it Accuracy} is the trainee's ability to keep alert during the inspection and thus develop the visual perception of the bridge condition. With situational awareness, inspectors can efficiently utilize the assistant drone in data collection and, later, effectively collaborate with machine learning algorithms in analyzing the inspection video data. The assessment module calculates the recall (the portion of the surface defects that the trainee detected correctly):
\begin{equation}
    \text{Rc}=X_{td}/X_d,
\end{equation}
and the precision (the portion of snapshots with a surface defect):
\begin{equation}
    \text{Pr}=X_{td}/X_{pd},
\end{equation}
to measure the accuracy. $\text{F}_\beta$ further integrates the recall and the precision as a single metric:
\begin{equation}
    \text{F}_\beta=\frac{(1+\beta^2)\text{Pr}\text{Rc}}{\beta^2\text{Pr}+\text{Rc}},
\end{equation}
where $\beta$ is a non-negative coefficient indicating the relative importance of recall with respect to precision. Setting $\beta$ as zero indicates precision is dominantly important, and setting it as $\infty$ means recall is dominantly important. $\beta$ is equal to one if precision and recall are equally important. $\text{F}_\beta$ is within [0, 100\%]. Accordingly, the score of accuracy is measured as 
\begin{equation}
P_A = \omega_f \text{F}_\beta.
\label{eq:accuracy}
\end{equation}
$\omega_f$ in Equation (\ref{eq:accuracy}) is set to be 100 and so the range of $P_A$ is [0, 100]. 

The trainee's scores on conformity, efficiency, safety, and accuracy are further standardized to be within the range from 0\% to 100\%. Then the standardized scores are presented as a Kiviat diagram to show the trainee's task performance on the four dimensions.

\subsubsection{Questionnaire-based Workload Assessment}
\label{subsubsec:Questionnaire}
After a simulation training is completed, the trainee is invited to fill out a questionnaire adopted from \cite{gabriel2016workload} and revised for TASBID. The questionnaire complements the objective assessment of TASBID. TABLE \ref{table:scales} lists the six aspects that the questionnaire asks. ``Time Pressure", ``Frustration", and ``In-task Feedback" are three aspects asked regarding the overall simulated inspection. ``Performance",  ``Mental Demand", and ``Physical Demand" are asked with respect to each phase or task of the inspection, including calibration, taking-off, individual tasks 1$\sim$4, and landing. Responses to questions are on a five-point likert scale: strongly agree (1), agree (2), neutral (3), disagree (4), and strongly disagree (5). ``Strongly agree" stands for the most positive response, and ``strongly disagree" stands for the most negative response. 
\begin{table}[htbp]
\centering
\caption{Self-Assessment Questionnaire}   
\small
\label{table:scales} 
\begin{tabular}{|m{0.4cm}|m{7.4cm}|}
%\begin{tabular}{cc}
\hline 
& \multicolumn{1}{c|}{\textbf{QUESTIONS}}\\
\hline 
\multirow{3}{*}[-1em]{\rotatebox{90}{Overall}} & \textbf{Time Pressure}: I finished the inspection without stress in regard of the required time.\\ \cline{2-2} 
%\hline 
 & \textbf{Frustration}: I never felt insecure, irritated, stressed, or discomforted during this task.\\\cline{2-2} 
%\hline 
 & \textbf{In-task Feedback}: The in-task feedback (e.g. battery level, speed, messages) were helpful for me.\\
\hline 
\multirow{3}{*}[-0.7em]{\rotatebox{90}{By tasks}} & \textbf{Performance}: I finished the task with a good performance.\\\cline{2-2} 
%\hline 
 & \textbf{Mental Demand}: It’s easy to finish the task.\\ \cline{2-2} 
%\hline 
 & \textbf{Physical Demand}: There was no physical activity (including pressing, pulling, turning, controlling, and holding) required in the task.\\ \cline{2-2}
\hline 
\end{tabular}
\end{table}

Heavy physical or mental demand may cause frustration and time pressure, and these psychological states may further influence the task performance. In-task feedback may mitigate the negative effect of the physical and mental loads posed on inspectors. The questionnaire can assist in causation analysis of the aforementioned relationship among causal factors (physical and mental demands), psychological states (time pressure and frustration), task performance, and the moderator (in-task feedback). 

\subsubsection{Repetitive Training for Improvement}
Practice using TASBID would help improve a trainee's task performance and the tolerance to physical and mental demands. The improvement is manifested by progressive changes in both performance measurements and subjective evaluation results. A hypothesis is that the post-study feedback would accelerate the learning of the trainee.

\section{Capability Demonstration}
\label{sec:Illustration}

A small-scale pilot study was conducted to demonstrate the functionality of TASBID. This study obtained the institutional IRB approval, which requires that participants are at least 18-years old and their participation is fully voluntary. 22 participants voluntarily contributed to the study. Among them, 4 are female, and 18 are male. Their ages are from 18 to 45, and their education backgrounds are Civil Engineering, Aerospace Engineering, Earth and Space Science, Physics, Computer Science and Engineering, and others. All participants have no prior experience with operating drones or serious games, but 10 out of 22 have the experience of playing video games for entertainment.

\subsection{Experiment Protocol}

\begin{figure}[htbp]
\centering
 \includegraphics[scale=0.3]{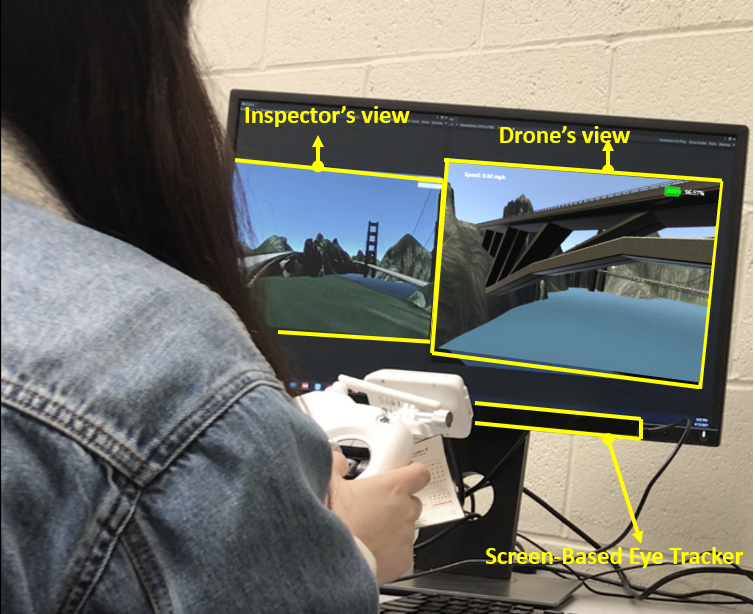}
\caption{A participant in the simulated inspection}
\label{fig:inspection_example}
\end{figure}

The experiment protocol for the simulation training is the following. In the beginning, an introduction to TASBID will be presented to the participant using a few PowerPoint slides. Then, a short tutorial \cite{simulation_TASBID} on the simulation training is presented as images and video clips with annotations. After that, the participant is offered an opportunity to practice the drone operation using the provided remote controller. The practice scene has some random variations from the scene for the simulation training. The simulated inspection starts after the participant feels she/he has enough practice and is ready for the study. After the training, the participant will fill out the questionnaire and then exit the study. The duration of the entire study can last 20$\sim$60 minutes, depending on the participant's prior experience with TASBID. Fig. \ref{fig:inspection_example} illustrates a participant operating the drone in the simulated inspection. 
Vivid videos of the inspection simulation can be found at the project website \cite{simulation_TASBID}.

\subsection{Performance in the Placement Training}
\begin{figure}[htbp]
\centering
 \includegraphics[scale=0.38]{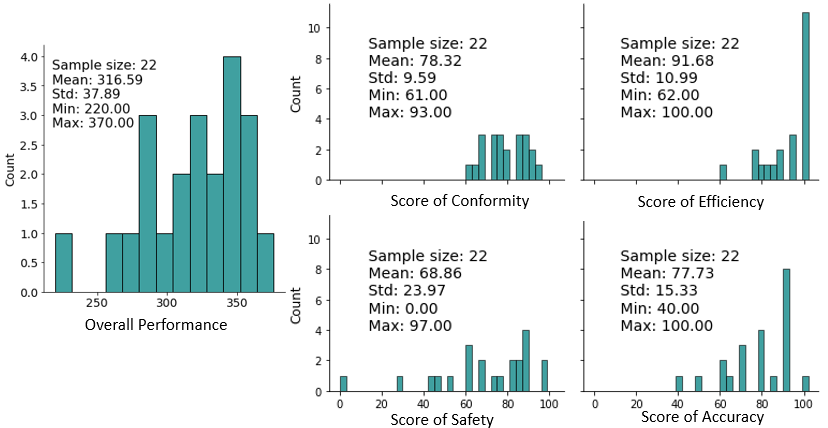}
\caption{Overall score distribution and marginal distributions on the four performance metrics}
\label{fig:performance}
\end{figure}

The task performance of a participant in the simulated inspection is calculated according to the assessment method presented in Section \ref{subsubsec:Task Performance}.The maximum overall score is 400, with 100 points allocated to each of the four performance metrics: conformity, efficiency, safety, and accuracy. The left chart in Fig. \ref{fig:performance} is the distribution of the 22 participants' overall scores in their first training. The chart indicates the heterogeneity in task performance. The participants’ overall score ranges from 220 to 370. The mean value is 316.59 and the distribution is skewed to the low end. The distribution of the overall score indicates a room for improvement. The four charts on the right of Fig. \ref{fig:performance} further show the score distributions on the four performance metrics, respectively. Efficiency has the largest mean (91.68) and the second smallest distribution range (38), indicating that it is the best achieved performance metric compared to others. Safety has the largest distribution range (97) but the smallest mean value (68.86), making it the most critical dimension for improvement. The mean scores of conformity (78.32) and accuracy (77.73) are well below the maximum 100, suggesting the need for improvement.

\begin{figure}[htbp]
\centering
 \includegraphics[scale=0.33]{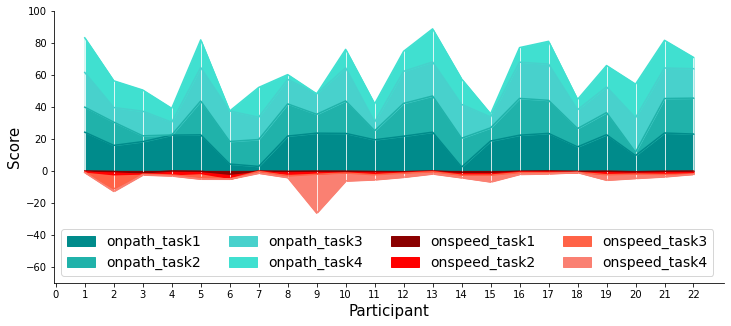}
\caption{Individuals’ conformity scores by tasks}
\label{fig:comformity}
\end{figure}

Fig. \ref{fig:comformity} further visualizes the conformity score of individual participants, broken down by their on-path and on-speed abilities in each of the four tasks. The figure shows that every participant has a unique conformity score profile in the first training; therefore, personalized feedback to individuals would be more helpful. For example, participant \#9 needs more practice for task 4 because of the low on-path score and the large loss due to speeding in that task. But this is not true for participant \#7 who needs to improve the on-path ability on task 1.

\begin{figure}[htbp]
\centering
 \includegraphics[scale=0.38]{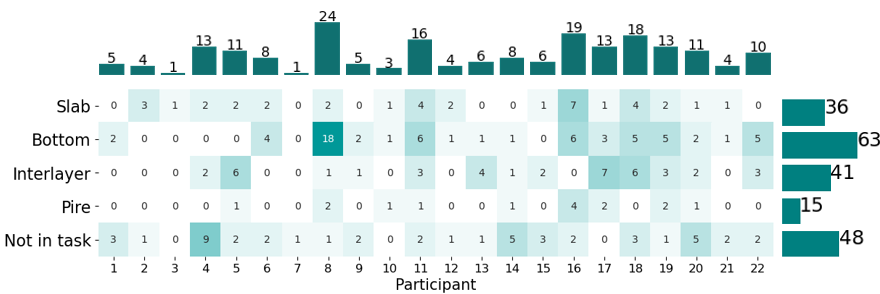}
\caption{Individuals’safety scores split by tasks}
\label{fig:Safety}
\end{figure}

During the first training, the 22 participants had 203 crashes in total. Fig. \ref{fig:Safety} counts the number of crashes by participants and tasks. The figure shows that participants’ ability to avoid crashes varies largely. Participants \#3 and \#7 each had only 1 crash, whereas participant \#8 had 24 crashes. The figure also indicates that the distribution of crashes on tasks varies largely from one participant to another. For example, participants \#4, \#17 and \#19 all had 13 crashes, but their safety concerns are different. Cumulatively, the proportion of crashes when inspecting the bridge bottom (63) is the largest, and the proportion when inspecting the pier (15) is the smallest.

\subsection{Self-assessment of the Placement Training}
\begin{figure}[htbp]
\centering
 \includegraphics[scale=0.58]{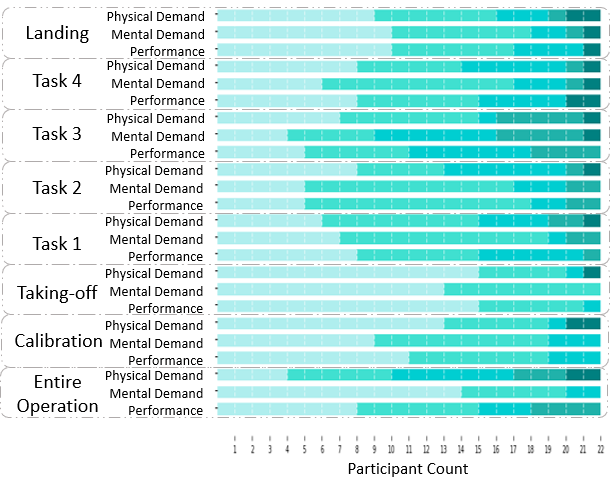}
\caption{The 22 participants’ responses to the post-study questionnaire in the first training}
\label{fig:survey_all}
\end{figure}

Fig.\ref{fig:survey_all} summarizes the distributions of the 22 participants’ responses to the questionnaire after they completed the first training. Only 45.5\% (10) participants agreed or strongly agreed that they were not frustrated by the job, and 68.2\% (15) participants agreed or strongly agreed that they did not feel time pressure in the job. But 90.1\% (20) participants agreed or strongly agreed that the in-task feedback is helpful. Operating a drone in a narrow space is likely to increase the mental demand. For example, the task of inspecting the bridge interlayer received the most negative answers compared to other tasks. Only 40.9\% (9) participants agreed or strongly agreed this task is low in mental demand, and only 50\% (11) participants agreed or strongly agreed that they performed well in this task. Operating a drone along the curved path with frequent position adjustments, like in task 2, seems to require more physical demand. Only 59.1\% (13) participants agreed or strongly agreed that task 2 is low in physical demand.

\subsection{Performance Improvement from the Repetitive Training}

\begin{figure}[htbp]
\centering
 \includegraphics[scale=0.33]{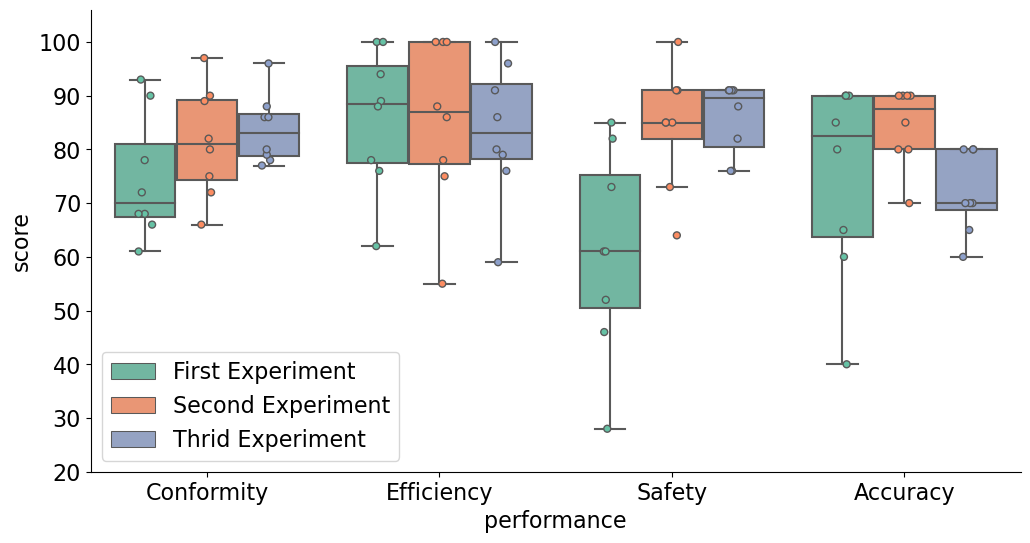}
\caption{Performance measurements from three times of training}
\label{fig:repeat_performances}
\end{figure}

Trainees can improve their skill of operating the assistant drone gradually through the repetitive training on TASBID. The post-study assessment result provided to participants may positively influence their learning outcome. For the illustration purpose, a focused group of 8 participants repeated the training for three times. The group was randomly drawn from the 22 participants, without referring to their placement training performance or other information.  Chi-squared homogeneity tests at the level of significance 0.05 confirm that the focused group can represent the 22 participants. The interval between two successive training sessions is at least 2 days. The overall inspection scene does not change over the repetitive training, but locations and size of surface defects are changed from one training to another. Fig. \ref{fig:repeat_performances} uses box plots to visualize the group’s performance achieved from the repetitive training. It is clear that, in the second or the third training, the group's average performance is improved on multiple performance metrics and the within-group variation was reduced.

TABLE. \ref{table:Ttests} further performed paired t tests on the mean increments of performance scores. Compared to the first training, the group improved the mean conformity score in the second training (p value of the upper tail test =0.018), and maintained the achieved improvement in the third training (p value of the two-tail test=0.312). The group’s mean improvement of the safety score after completing the second training was significant (p value of the upper tail test=0.006), and the improvement was maintained in the third training (p value of the two-tail test=0.649). The improvements of conformity and safety in the second and third training did not worsen the time efficiency. The group maintained the efficiency throughout the three times of training (p values of two-tail tests $\geq$ 0.380). The group improved the mean accuracy after completing the second training (p value of the upper tail test = 0.048). But the mean accuracy was reduced after finishing the third training (p value of the lower tail test = 0.000), mainly due to the increased difficulty to visually detect surface defects.

\begin{table}[htbp]
\begin{threeparttable}
\footnotesize
\centering
\caption{Paired t tests of the learning effect} 
\label{table:Ttests} 
\begin{tabular}{l|ccc|ccc}
    \toprule
    \multirow{2}{*}{ } &
      \multicolumn{3}{c|}{\textbf{Conformity}} &
      \multicolumn{3}{c}{\textbf{Efficiency}} \\
      & {1 vs. 2 } & {1 vs. 3}& {2 vs. 3} & {1 vs. 2} & {1 vs. 3}& {2 vs. 3}  \\
      \midrule
    t value & 2.609 & 2.969 & 1.090 & -0.175  & -0.938  & -0.661  \\
    p value & 0.018$^u$  & 0.010$^u$  & 0.312$^t$  & 0.866$^t$  & 0.380$^t$  & 0.530$^t$ \\
     \toprule
     \multirow{2}{*}{ } &
      \multicolumn{3}{c|}{\textbf{Safety}} &
      \multicolumn{3}{c}{\textbf{Accuracy}}\\
      & {1 vs. 2 } & {1 vs. 3}& {2 vs. 3} & {1 vs. 2} & {1 vs. 3}& {2 vs. 3}  \\
      \midrule
    t value & 3.340  & 3.498 & 0.475 & 1.930 & -0.662 & -7.638 \\
    p value &  0.006$^u$  & 0.005$^u$  & 0.649$^t$  &  0.048$^u$ & 0.529$^t$ & 0.000$^l$\\
        \bottomrule
  \end{tabular}
    \begin{tablenotes}
        \scriptsize
         \item Note:
            \item $\cdot$ ``b" vs. ``a": the increment tested is the score in ``a" minus the score in ``b".
            \item $\cdot$ the superscripts ``$u$", ``$l$", and ``$t$" indicate the upper-tail test, lower-tail test, and two-tail test, respectively.
            %\item $\cdot$ The null hypothesis for "1 v 2 (or 3)" is "$2^{nd}$ (or $3^{rd}$) Score  - $1^{st}$ Score $\leq$ 0", the alternative hypothesis is "$2^{nd}$ (or $3^{rd}$) Score  - $1^{st}$ Score $\geq$ 0"
            %\item $\cdot$ The null hypothesis for "2 v 3" is "$3^{rd}$ Score  - $2^{nd}$ Score = 0", the alternative hypothesis is "$3^{rd}$ Score  - $2^{nd}$ Score $\neq$  0"
    \end{tablenotes}
 \end{threeparttable}
\end{table}

\subsection{Self-assessment of the Repetitive Training}

\begin{figure}[htbp]
\centering
 \includegraphics[width=\columnwidth]{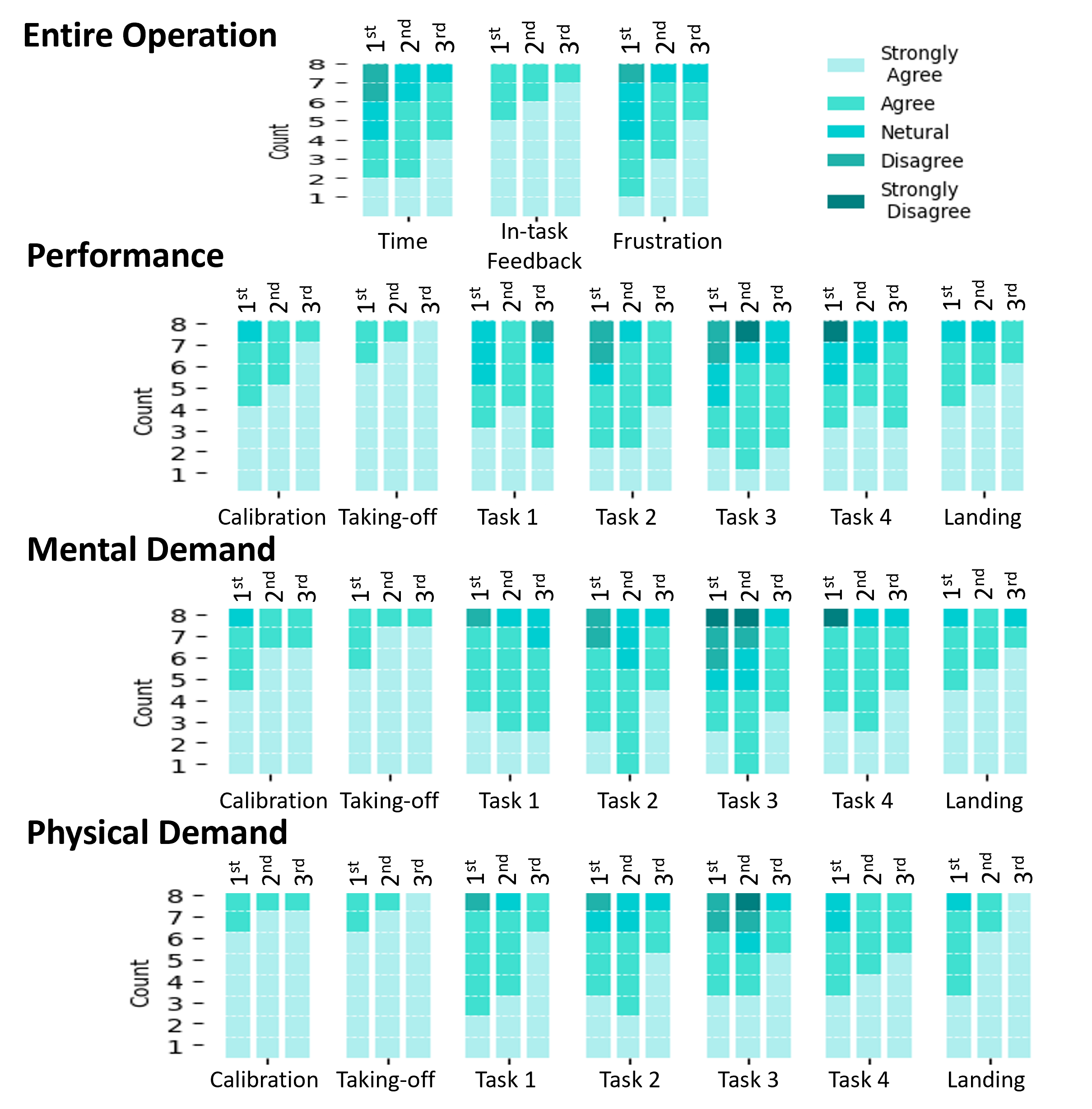}
\caption{Participants’ responses to  the  post-study  questionnaires in the repetitive training}
\label{fig:survey_repeat}
\end{figure}

The repetitive training helps trainees improve not only their task performance, but confidence and comfort in operating an assist drone for bridge inspection.  Fig.\ref{fig:survey_repeat} summarizes the self-assessment of the eight participants after finishing each training. The figure implies that participants struggle more when inspecting the bridge from a narrow space (task 2) or on curved paths (tasks 3 and 4). But, overall, the response to the questionnaire turns to be more positive after they practiced the inspection using BASBID.

\subsection{Performance Analysis for Individual Trainees}
\begin{figure}[htbp]
\centering
 \includegraphics[scale=0.43]{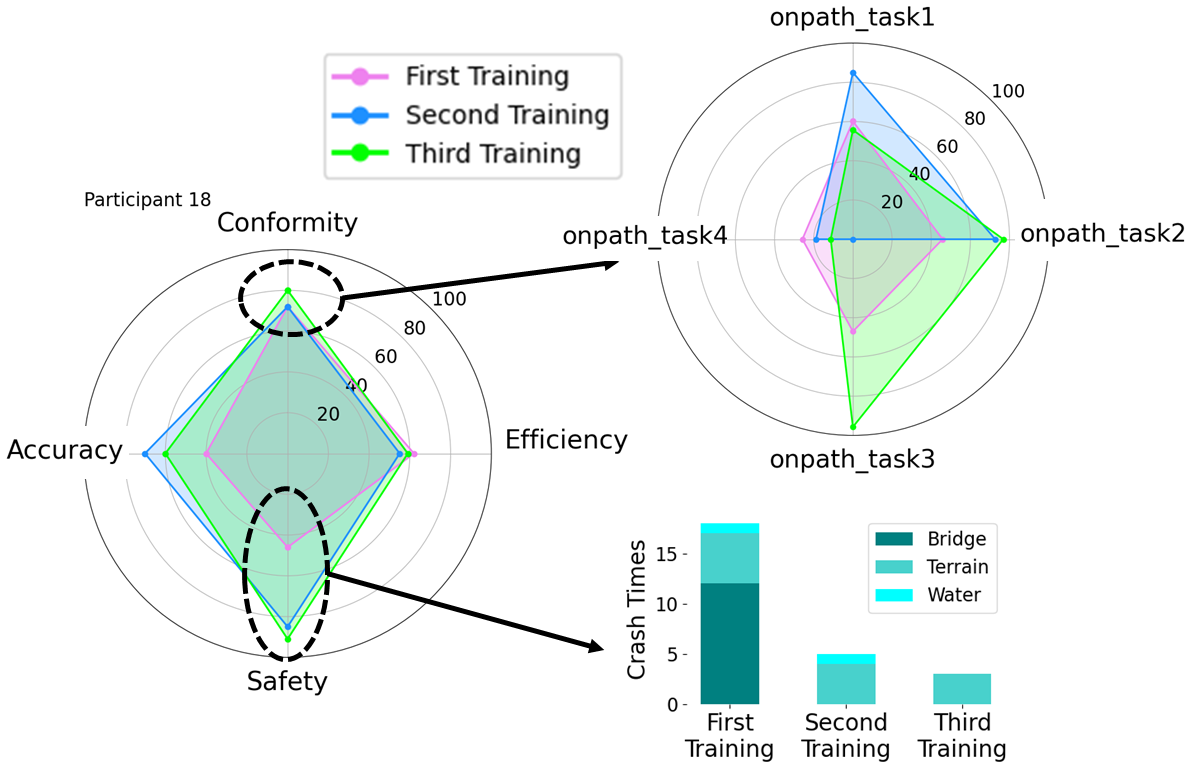}
\caption{Performance of participant \#18}
\label{fig:participant 18}
\end{figure}

TASBID can determine the specific strengths and weaknesses for any trainee, identify causes of the weaknesses, and track the training progress. Fig. \ref{fig:participant 18} presents participant \#18's performance in the three times of training as an illustrative example. The participant improved the accuracy score from 40 to 70 in the second training. Although the accuracy score dropped to 60 in the third training, that change was mainly caused by the increased challenge in recognizing the cracks visually in that experiment. The efficiency score of participant \#18 did not change much, ranging from 22.5 to 24 minutes in the repetitive training. The participant improved the conformity score in the third training, from 72 to 80. The upper-right figure further shows the on-path scores of the participant in performing each of the four tasks. In the second training, the participant improved the on-path scores on tasks 1 and 2, but she/he did not perform task 3 due to insufficient time. In the third training, the participant significantly improved the on-path score for task 3. Overall, the participant needs more practice to improve the ability to fly the drone along reference paths. The participant clearly improved her/his safety score in the second training and maintained the safety performance in the third training. The plot at the bottom-right indicates that the participants crashed into the bridge twelve times in the first training, but not at all in the second and third training. The number of crashes to the terrain or the water body has a decreasing trend over the three times of training.

\subsection{Subjective vs. Objective Evaluations}
Overall, this pilot study shows a consistency between the subjective evaluation result and the objective assessment result. But self-assessment is subject to a certain degree of bias, which may lead to inconsistent results \cite{donaldson2002understanding,dienes2004assumptions}. The self-ratings of task performance by a few trainees seem to contradict their actual performance. For example, participant \#8 made 28 out of 100 points on safety from the first training due to many crashes. Although the participant should know that (because a warning message is shown on the screen if a crash happened), the participant strongly agreed that she/he performed well in the tasks. Participant \#13 made 100 out of 100 on safety and 97 out of 100 on conformity, but the participant did not strongly agree that she performed well. Biases are present in their responses to another question ``I finished the inspection without stress regarding the required time". Two striking contrasts are the answers from participants \#10 and \#18. Participant \#10 spent 13.75 minutes to finish all four tasks, but her/his response to this statement is a disagree. Participant \#18 spent 24 minutes completing three tasks only, but the answer is neutral. Another example of the contradictory response is from participant \#2. This participant kept on the reference path for inspecting the bridge interlayer for about 30\% of the task time, but the participant believes she/he performed well in this task. The relatively good performances on some dimensions (i.e., quick completion and few crashes) probably made the participant underestimate the consequence of the insufficient data collection. From the aforementioned contradictory examples, the pilot study supports the use of objective assessment. The post-study analysis can tell what happened by analyzing the captured training data to provide objective feedback to the trainee.

\section{Conclusions and Future Work}
\label{sec:Conclusion}

This paper designed and developed a virtual reality-based training and assessment system named TASBID for bridge inspectors collaborating with an assistant drone to collect data at inspection sites. The pilot study, although is in a small-scale, demonstrated that TASBID can objectively identify the training needs of individuals in detail and further help them develop the skill and confidence in collaborating with a drone in bridge inspection. This study shares the source code with the public. Prospective users can easily revise it to adapt to their own specific studies or needs \cite{TASBID_Github}.

The training and assessment introduced in this paper have built a foundation for adding the semi-autonomous mode to TASBID. With the semi-autonomous mode, the drone will fly automatically, but the inspector can disengage the autonomous mode of the drone and take control of it when needed. Besides, a gap is present between the simulation created in Unity and the real-world inspection scene. A generative adversarial network can convert the simulation to a more realistic scene, thus providing improved visual stimuli to inspectors.  Furthermore, this paper focuses on the system design and development, thus only conducting a small-size pilot study to demonstrate the system functionality. Factorial experiments at a larger scale would be necessary for comprehensive system testing and improvement. TASBID can integrate a multi-modal biometric sensor system comprised of an eye tracker, electromyography, and inertial measurement units. Deep neural networks need to be developed for analyzing the biometric sensor data to reliably detect and classify human states and for creating other methods of human-drone interactions. This paper has built a foundation for exploring the above-discussed opportunities.

\section*{Acknowledgement}
All authors received financial support from National Science Foundation through the grants ECCS-\#2026357 and CMMI-\#1646162.

\bibliographystyle{IEEEtran}
\bibliography{references.bib}

\end{document}